\documentclass[letterpaper, 10 pt, conference]{ieeeconf}  

\IEEEoverridecommandlockouts                              

\usepackage{graphics} 
\usepackage{epsfig} 
\usepackage{mathptmx} 
\usepackage{times} 
\usepackage{amsmath} 
\usepackage{amssymb}  

\usepackage{graphicx}
\usepackage{booktabs}

\usepackage{bm}
\usepackage{bbm}
\usepackage{mathtools}

\usepackage{textcomp}
\usepackage{xcolor}
\usepackage{wrapfig}
\usepackage{cite}
\usepackage{algorithm}
\usepackage{algorithmic}
\usepackage{dsfont}
\usepackage{multirow}
\usepackage{tabularx}

\usepackage{subcaption}
\usepackage[pagebackref,breaklinks,colorlinks]{hyperref}

\usepackage[capitalize]{cleveref}
\crefname{section}{Sec.}{Secs.}
\Crefname{section}{Section}{Sections}
\Crefname{table}{Table}{Tables}
\crefname{table}{Tab.}{Tabs.}

\DeclareMathAlphabet{\mathcal}{OMS}{cmsy}{m}{n}
\newcommand{\Method}{SeaSplat\,}

\newcommand{\rbt}[2]{{T^{#1}_{#2}}} 

\newcommand{\loss}{\mathcal{L}}     

\newcommand{\link}[1]{\textcolor{magenta}{\href{#1}{#1}}}

\title{\LARGE \bf
SeaSplat: Representing Underwater Scenes with 3D Gaussian Splatting and a Physically Grounded Image Formation Model
}

\author{Daniel Yang$^{1, 2}$, John J. Leonard$^{1}$, and Yogesh Girdhar$^{2}$
\thanks{$^{1}$Computer Science and Artificial Intelligence Laboratory, Massachusetts Institute of Technology, Cambridge, MA, USA \texttt{\{dxyang, jleonard\}@mit.edu}}%
\thanks{$^{2}$Applied Ocean Physics \& Engineering Department, Woods Hole Oceanographic Institution, Woods Hole, MA, USA \texttt{\{dayang, yogi\}@whoi.edu}}%
}

\begin{document}

\maketitle
\thispagestyle{empty}
\pagestyle{empty}

\begin{abstract}

We introduce \Method, a method to enable real-time rendering of underwater scenes leveraging recent advances in 3D radiance fields. Underwater scenes are challenging visual environments, as rendering through a medium such as water introduces both range and color dependent effects on image capture. We constrain 3D Gaussian Splatting (3DGS), a recent advance in radiance fields enabling rapid training and real-time rendering of full 3D scenes, with a physically grounded underwater image formation model. Applying \Method to the real-world scenes from SeaThru-NeRF dataset, a scene collected by an underwater vehicle in the US Virgin Islands, and simulation-degraded real-world scenes, not only do we see increased quantitative performance on rendering novel viewpoints from the scene with the medium present, but are also able to recover the underlying true color of the scene and restore renders to be without the presence of the intervening medium. We show that the underwater image formation helps learn scene structure, with better depth maps, as well as show that our improvements maintain the significant computational improvements afforded by leveraging a 3D Gaussian representation. Code, data, and visualizations are available at \link{https://seasplat.github.io}

\end{abstract}

\section{Introduction}
\label{sec:intro}

Underwater vehicles equipped with imaging capabilities have increasingly been deployed for complex and adaptive tasks including environmental monitoring \cite{manderson2017robotic, girdhar2023curee, mccammon2024discovering}, visual target tracking \cite{cai2023semi, katija2021visual}, and exploring underwater environments \cite{joshi2022underwater, johnson2017high}. However, visual perception underwater presents a significantly different set of characteristics than images collected in air, making it challenging to directly utilize many advances in computer vision to this domain \cite{islam2020fast, de2021impact}. 

Images collected underwater exhibit degradation because of the inherent effects of imaging through a medium, water. Light propagating through water is affected by range-dependent and spectrally-sensitive attenuation as well as backscatter. In practice, the attenuation causes underwater images to be lacking in certain colors, specifically red, while having an excess of others, blue and green, while the backscatter causes underwater images to exhibit a veiling or hazy effect. The magnitude of these effects varies with the range of what's being imaged from the camera as well as the depth from the sea surface. 

\begin{figure}[ht]
    \centering
    \includegraphics[width=0.9\columnwidth]{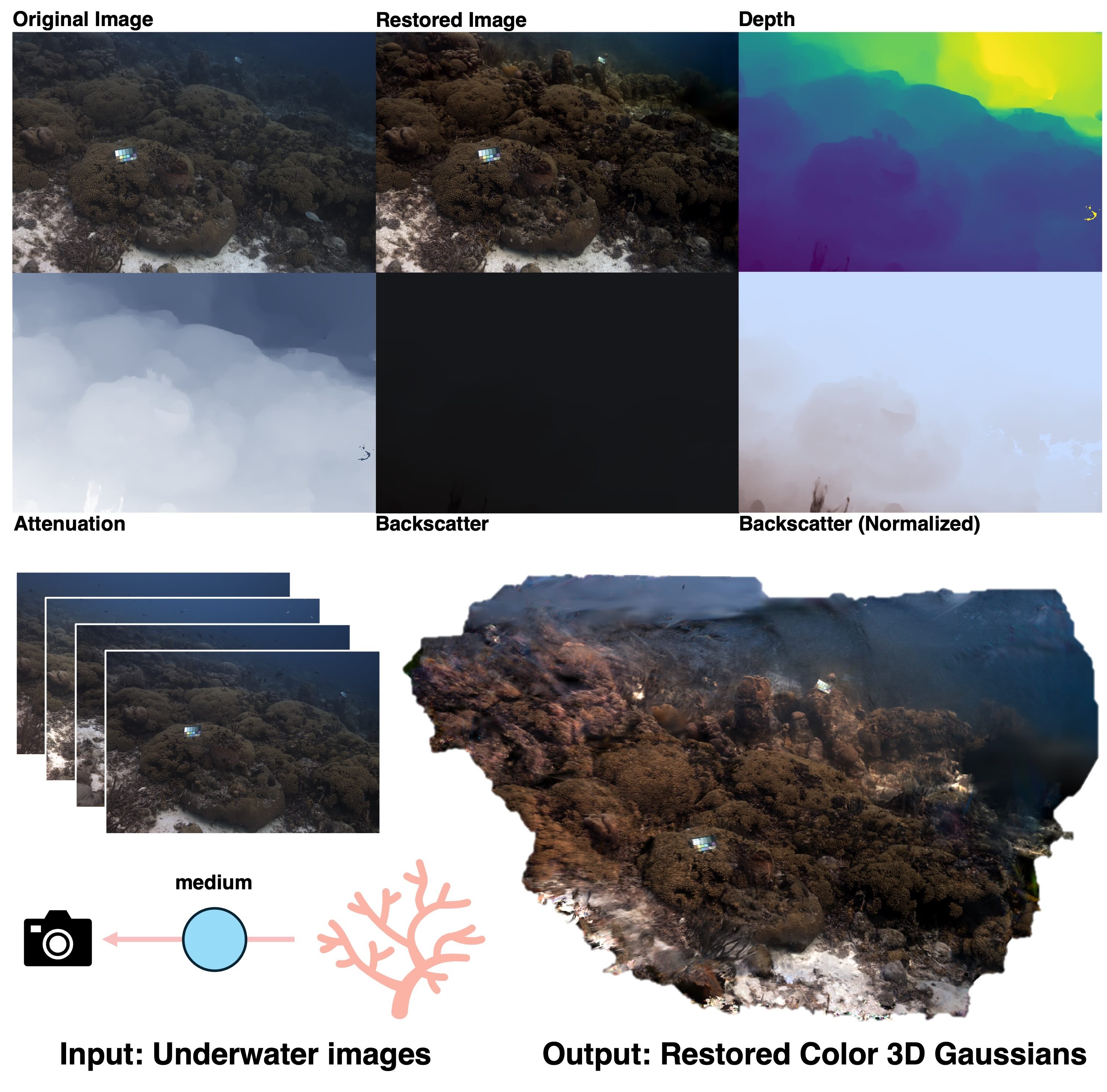}
    \caption{\textbf{Overview} \Method constrains a 3D Gaussian representation with a physics-based underwater image formation model to restore the scene against effects caused by imaging through water and provide higher quality novel view synthesis. \Method estimates medium parameters for the depth and wavelength dependent effects of backscatter and attenuation, while also enabling rapid view synthesis of the scene with and without the medium.}
    \label{fig:teaser}
    \vspace{-0.5cm}
\end{figure}

Radiance field methods, catalyzed by the use of neural networks to parameterize volumetric density and color by NeRF \cite{mildenhall2020nerf}, have led to significant advances in reconstructing 3D scenes and achieving high-quality novel-view synthesis. More recently, 3D Gaussian Splatting \cite{kerbl3Dgaussians} has seen success in enabling real-time, photorealistic rendering of these 3D environments. However, these methods typically assume atmospheric (e.g., imaging through air) conditions for the data where color is relatively constant with distance, leading to undesirable effects such as floaters and other geometric inconsistencies when applied to underwater environments. 
We propose \Method, a method that combines 3D Gaussian Splatting with a physically grounded underwater image formation model. By simultaneously learning the parameters of the medium along with the underlying 3D representation, we can restore the scene's true color while more accurately estimating the scene's geometry, as shown through higher fidelity and geometrically consistent rendering of novel viewpoints. We highlight our method's ability to estimate the true color of the scene on real world underwater data collected by divers in various coral environments, on simulated scenes in outdoor environments, as well as on data collected by an autonomous underwater vehicle. 

\section{Related Works}
\label{sec:related-works}

\subsection{Radiance fields}
Neural radiance fields (NeRF) \cite{mildenhall2020nerf} have spurred significant advances in 3D reconstruction and novel view synthesis. NeRF encodes the volumetric density and color of a scene implicitly within a multilayer perceptron (MLP). For rendering a viewpoint, dense samples of the volume, obtained through sampling along rays going through each image coordinate, are used to query the MLP before being aggregated along each ray. A photometric error loss can be applied on this differentiably rendered image to learn the weights of the MLP. This procedure is computationally expensive, especially in the dense sampling and querying of the MLP. A flurry of extensions of NeRFs have shown success in improving both their quality and speed \cite{fridovich2022plenoxels, muller2022instant, park2021nerfies, barron2021mipnerf, barron2022mipnerf360}. For a more in-depth review of these methods, we refer the reader to \cite{tewari2022advances} and \cite{xie2022neural}. In contrast to this paradigm, a recent advance, 3D Gaussian Splatting \cite{kerbl3Dgaussians}, explicitly parameterizes the radiance field with 3D Gaussians which can be efficiently rendered via rasterization while maintaining state-of-the-art visual quality. 

\subsection{Underwater color restoration}

Color correction of underwater imagery \cite{bryson2016true, carlevaris2010initial} has relied upon dehazing approaches for above-ground imagery \cite{he2009haze, berman2016non}. Dehazing approaches try to decompose an observed image into the scene radiance, atmospheric light, and the transmission, a term exponentially decaying with depth and governed by a wavelength-independent attenuation coefficient. However, \cite{akkaynak2018revised} show that such a model is insufficient for the underwater domain. Instead, they propose a revised model where the attenuation coefficient is wavelength dependent, e.g. varies by color channel, and where the transmission coefficients governing attenuation and the atmospheric light are different. However, even with a more complex model that captures the nuances of underwater image formation, estimation of the parameters have been challenging. With a color chart present at multiple distances, one can estimate the parameters of the model \cite{akkaynak2018revised}. Further work has lead to parameter estimation when depth information is present \cite{akkaynak2019seathru}. 

Instead of model fitting, others perform color correction using learning-based methods \cite{fabbri2018enhancing, li2017watergan}. These approaches frame the problem as learning a mapping between two distributions of data with generative adversarial networks \cite{goodfellow2014generative}. However, these methods do not guarantee consistency in the medium effects across a scene or in varying environmental conditions from the data used to train the model. 

More recent work combines underwater color restoration with neural radiance fields. SeaThru-NeRF \cite{levy2023seathrunerf} integrates the underwater image formation model into a neural radiance field framework \cite{barron2022mipnerf360}. They augment NeRFs by sampling medium parameters per viewing direction, instead of densely per ray as typically done, and achieve both color correction of images as well as novel-view synthesis. Another work \cite{zhang2023beyond} similarly integrates an underwater image formation model into a neural radiance field framework and learns physical properties of the scene like albedo. However, \cite{zhang2023beyond} focuses on backscatter effects resulting from a light source moving with the camera and is better suited for capturing targeted objects or small areas underwater as opposed to the scene-scale reconstructions of SeaThru-NeRF \cite{levy2023seathrunerf}.
\section{Preliminaries}
\label{sec:prelim}

\subsection{3D Gaussian Splatting}
\label{sec:prelim-gs}
3D Gaussian Splatting (3DGS) \cite{kerbl3Dgaussians} parameterizes a scene with a set of 3D Gaussians, each with a mean $\bm{\mu}$, covariance $\bm{\Sigma}$ derived from a scale $\bm{S}$ and rotation $\bm{R}$, opacity $\bm{o}$, and color $\bm{c}$ derived from spherical harmonic coefficients. Given a viewpoint, $\bm{T^{\text{cam}}_{\text{world}}}$, we can efficiently render an image by sorting the Gaussians from front to back, projecting them onto the camera plane, and alpha-compositing the splatted Gaussians. For a pixel $\bm{x}$, its color can be computed as: 

\vspace{-0.3cm}
\begin{equation}
    \label{eq:gs-color}
    C(x) = \sum_{i =1}^N c_i\alpha_i(x)\prod_{j=1}^{i-1}(1-\alpha_j(x))
\end{equation}

\noindent where $c_i$ and $\alpha_i$ are the color and density of the $\textit{i}^{th}$ 3D Gaussian. Such a representation is optimized with an L1 reconstruction loss combined with a D-SSIM term:

\vspace{-0.3cm}
\begin{equation}
    \label{eq:gs-vanilla-loss}
    \mathcal{L}_{\text{original}} = (1 - \lambda) \mathcal{L}_1 + \lambda \mathcal{L}_{\text{D-SSIM}}
\end{equation}

\subsection{Underwater Image Formation}
\label{sec:prelim-uwimage}
Underwater, two wavelength and distance dependent processes affect the propagation of light and how images are captured. First, as light propagates through water, its intensity is \textit{attenuated} in a spectrally-selective manner, with red light being attenuated much more than blue light over similar distances \cite{pegau1997absorption}. Second, particles in the water column reflect light from various sources towards the camera leading to \textit{backscatter}, also referred to as veiling light. Both effects vary with range between the camera and the imaged object - as the object is further, light reflected by the object towards the camera is more attenuated and more particles in the water column between the object and camera lead to more backscatter. 

These processes can be modelled as the following \cite{akkaynak2018revised}:

\vspace{-0.3cm}
\begin{equation}
    \label{eq:uw-image-formation}
    I = J \cdot e^{-\beta^D \cdot Z} + B^\infty \cdot (1 - e^{-\beta^B \cdot Z}),
\end{equation}

\noindent where $I$ is the captured image, $J$ is the underlying color that would be captured if there was no medium, $\beta^D$ and $\beta^B$ are the attenuation and backscatter coefficients respectively, $B^\infty$ is the backscatter water color at infinity, and $Z$ is the depth image. While depth from the surface also plays a role in the amount of visible light, note that depth here refers to distance or range from the camera. The attenuated true-color image, $J \cdot e^{-\beta^D \cdot Z}$, is also referred to as the direct image, $D$.
\begin{figure*}[tp]
    \centering
    \includegraphics[width=\linewidth]{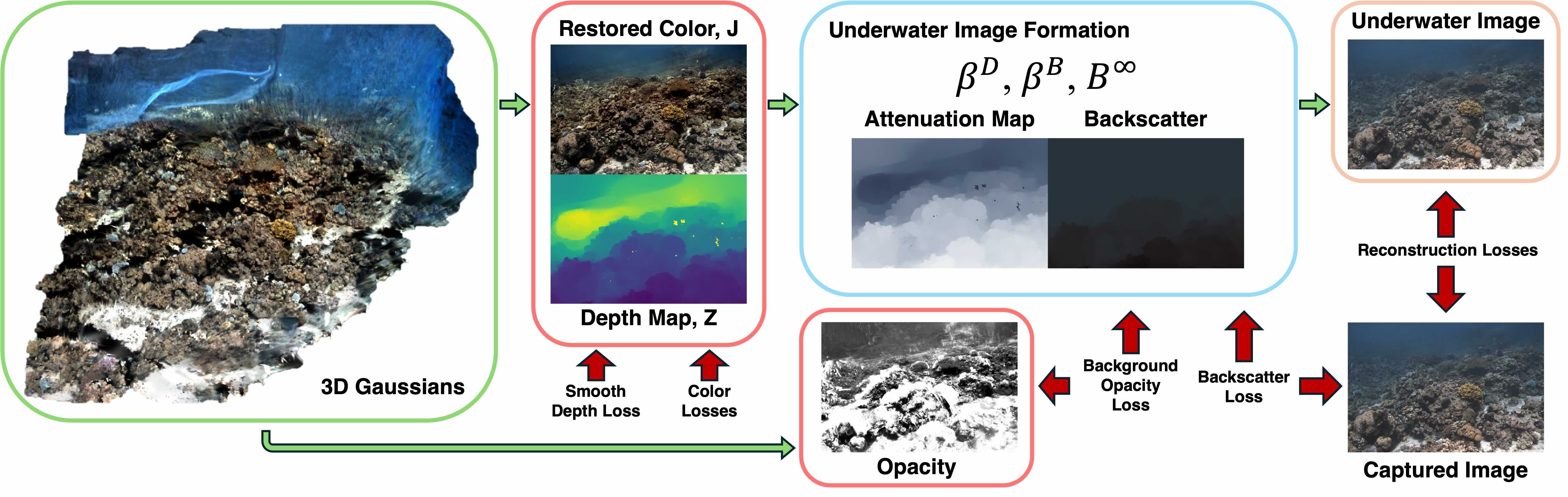}
    \vspace{-0.5cm}
    \caption{\textbf{SeaSplat} We constrain 3D Gaussian Splatting \cite{kerbl3Dgaussians} with the underwater image formation model as shown here, requiring our underlying 3D Gaussians to model the true, restored color of the scene thus accounting for the effects of rendering through a medium, water. Combining the channel-specific backscatter and attenuation parameters with the depth map, we can estimate the effects of backscatter and attenuation at the specified viewpoint. These effects can then be combined with the estimated true color to generate an in-medium, underwater image where we can apply standard photometric losses with the captured image. Losses are applied at different steps of this model to help learn the parameters of the underlying 3D Gaussian representation as well as the underwater image formation parameters.}
    \label{fig:overview}
    \vspace{-0.5cm}
\end{figure*}

\section{SeaSplat: Underwater 3D Gaussians}
\label{sec:method}

\subsection{Our approach}

We present \Method, a method based on 3D Gaussians enabling novel view synthesis while simultaneously estimating the underlying restored color of the environment, as if there was no medium (water) present. Given a set of images $I \in \mathbb{R}^{H\times W\times 3}$, the camera intrinsics $K \in \mathbb{R}^{3\times 3}$, and the world-to-camera extrinsics $\rbt{cam}{world} \in \mathbf{SE}(3)$ of each frame obtained through structure from motion (e.g. COLMAP \cite{schoenberger2016sfm}), we aim to learn a 3D Gaussian representation of the underlying, color-corrected scene while also learning the parameters of the medium effects, backscatter and attenuation. An overview of this model is shown in \cref{fig:overview}.

By modelling the color-corrected scene, we can estimate the depth and wavelength dependent effects with the depth from our 3D Gaussian representation and the physical model described in \cref{eq:uw-image-formation}. We can then combine these effects with the underlying color. Intuitively, constraining the 3D Gaussian representation with the physical model pushes the 3D Gaussians to be better situated in 3D space with more physically plausible depth.

Concretely, for a given camera viewpoint we can render the underlying true color image $\hat{J}$ as well as the estimated depth map $\hat{Z}$ of the 3D scene. Using the image formation model, we can calculate the attenuation map for each channel, $\hat{A} = e^{-\beta^D \cdot \hat{Z}}$, as well as backscatter image, $\hat{B} = B^\infty \cdot (1 - e^{-\beta^B \cdot \hat{Z}})$. Finally, our reconstructed underwater image is simply $\hat{I} = \hat{J} \cdot \hat{A} + \hat{B}$.

\subsection{Optimization}
Though our 3D Gaussian representation is constrained to satisfy the underwater image formation model, using solely the standard loss of \cite{kerbl3Dgaussians} is insufficient to lead to meaningful, non-trivial solutions. For example, a model that adds no backscatter and performs no attenuation could still satisfy the optimization objective. Likewise, as we are learning both the depth and these medium parameters, the optimization could also push the depth to be implausible given a poor estimate of the medium parameters. Thus, we introduce additional loss constraints while also interleaving optimization of the medium parameters with optimization of the underlying 3D Gaussian representation.

To optimize the backscatter parameters, we adopt a backscatter loss developed by DeepSeeColor \cite{jamieson2023deepsee}, a variant of the dark channel prior loss for haze removal \cite{he2009haze}:

\vspace{-0.3cm}
\begin{equation}
    \label{eq:loss-backscatter}
    \begin{aligned}
        \loss_\text{bs} = \sum_{(i,j)} \sum_{c} \big(\max\{\hat{D}_c(i,j), 0\} + k \min\{\hat{D}_c(i,j), 0\}\big),
    \end{aligned}
\end{equation}

\noindent with hyperparameter $k>1$, the direct image $\hat{D}$ as the backscatter-removed observed image, $\hat{D} \coloneqq I - \hat{B}$, and $\hat{D}_c$ as the direct image in a specific color channel, $c \in \{r,g,b\}$. $\hat{B}$ is calculated using the estimated depth from the 3D Gaussian representation, $\hat{Z}$, which is detached to prevent gradients from flowing through. Thus, gradients do not affect the underlying 3D Gaussian representation, but the learned backscatter parameters do constrain the image formation model used in other losses. 

Similar to other underwater color correction works \cite{nathan2024osmosis, jamieson2023deepsee}, we encourage the average values of each color channel to approach the middle of the color range, the gray world hypothesis \cite{buchsbaum1980spatial}, and penalize color oversaturation:

\vspace{-0.3cm}
\begin{align}
    \loss_\text{gw} &= \frac{1}{3} \sum_{c}
        \Bigg(
            \Big( 
                \frac{1}{N} \sum_{i,j} \hat{J}_c(i,j)
            \Big) - 0.5
        \Bigg)^2 
    \label{eq:loss-gw} \\
    \loss_\text{sat} &= \sum_{i,j} \sum_{c} \max\big( \hat{J}_c(i,j) - T_{sat}, 0\big) 
    \label{eq:loss-sat},
\end{align}
\vspace{-0.1cm}

\noindent with $N$ as the number of pixels in $\hat{J}$ and $T_{sat}$ as a threshold empirically set to 0.7.

Likewise, to emphasize the recovery of details far away, where backscatter and attenuation may have the largest effect on visual appearance, we add a depth-weighted reconstruction loss, without allowing gradients to flow through the depth image. This is in addition to the original reconstruction loss of 3DGS \cite{kerbl3Dgaussians}.

\vspace{-0.3cm}
\begin{equation}
    \label{eq:loss-depthweightedrecon}
    \loss_{\text{Z-recon}} = || Z_\text{detach} \cdot (I - \hat{I}) ||_1
\end{equation}

We regularize the estimated depth with an edge-aware total variation loss \cite{godard2017unsupervised}, pushing the depth map to be smooth while allowing for depth to vary more along areas of high gradient in the observed camera image (e.g. edges).

\vspace{-0.3cm}
\begin{equation}
    \label{eq:loss-depthsmooth}
    \loss_{Z_\text{smooth}} = \sum_{i,j}(e^{-\nabla_x I}|\nabla_x \hat{Z}| + e^{-\nabla_y I}|\nabla_y \hat{Z}|)
\end{equation}

Because of the dominance of the reconstruction loss in guiding optimization of the 3D Gaussian representation, their often are Gaussians with high opacity added in regions that do not necessarily represent objects underwater, for example the water column. To counteract this, we add a loss where pixels with color similar in Euclidean distance to the estimated $B^\infty$ color are pushed to have a low alpha mask value.

\vspace{-0.3cm}
\begin{equation}
    \label{eq:loss-binf-background}
    \loss_{\text{op}} = \sum_{i,j} \alpha(i,j) \cdot \mathds{1}_{|| I(i,j) - B^\infty ||^2_2 < T_{sim}}
\end{equation}

Our total loss becomes the following, with $\loss_{\text{GS}}$ corresponding to the original 3D Gaussian Splatting \cite{kerbl3Dgaussians} loss: 

\vspace{-0.3cm}
\begin{equation}
    \label{eq:splash-loss}
    \mathcal{L} = \mathcal{L}_{\text{GS}} + \mathcal{L}_{\text{bs}} + \mathcal{L}_{\text{gw}} + \mathcal{L}_{\text{sat}} +  \mathcal{L}_{\text{op}} + \mathcal{L}_{Z_\text{smooth}} + \mathcal{L}_{\text{Z-recon}}
\end{equation}

\subsection{Implementation}
We build upon the code from 3D Gaussian Splatting \cite{kerbl3Dgaussians} with modifications to enable differentiably render depth maps and alpha masks \cite{wewer24latentsplat, turkulainen2024dnsplatter}. Depth rendering is implemented as a second rendering pass with the colors passed into the rendering function replaced with the z value of each 3D Gaussian. We use zero order spherical harmonics, such that the color of each Gaussian has no view dependencies. 

Similar to \cite{jamieson2023deepsee}, attenuation and backscatter coefficients, $\beta^D$ and $\beta^B$ are each implemented as a (1, 1, 1, 3) kernel that can be efficiently applied to the depth image, $D$, through convolution. $B^\infty$ is implemented as a differentiable parameter.

\section{Experiments}
\label{sec:experiments}

\subsection{Experimental Setup}

\paragraph{Datasets}
We use the multi-view underwater scenes published with Sea-ThruNeRF \cite{levy2023seathrunerf} consisting of four scenes with COLMAP \cite{schoenberger2016sfm} extracted camera poses --- two in the Red Sea (Japanese Gardens and IUI3), one in the Carribean Sea (Curaçao), and one in the Pacific Sea (Panama). These scenes cover a range of geographical locations and water conditions. RAW images were acquired with a DSLR in an underwater housing with a dome port and white-balanced.

We also use an underwater scene consisting of a color chart on the seafloor of Salt Pond Bay in the US Virgin Islands collected by a remote controlled underwater robot with downward facing cameras, CUREE \cite{girdhar2023curee}. The robot moved between low and high altitudes to highlight the image degradation with increased distance. Notably, this dataset features downward facing imagery of the seafloor as opposed to the forward facing imagery containing the water column used in the SeaThru dataset. COLMAP \cite{schoenberger2016sfm} was also used to extract camera poses.

We also generate a synthetic, simulated dataset based off the outdoor \textit{garden} scene from Mip-NeRF 360 \cite{barron2022mipnerf360}. We estimate depth maps using 3DGS \cite{kerbl3Dgaussians} and simulate a foggy and underwater scene, similar to \cite{levy2023seathrunerf}. Water is added based off \cref{eq:uw-image-formation} with $\beta_D = [2.6, 2.4, 1.8]$, $\beta_B = [1.9, 1.7, 1.4]$, and $B^\infty = [0.07, 0.2, 0.39]$. Fog is added with $\beta_D = \beta_B = 2.4$, dropping the wavelength dependencies and the difference between backscatter and attenuation terms \cite{berman2016non, he2009haze}.

\paragraph{Comparisons}
Closest to our work is SeaThru-NeRF \cite{levy2023seathrunerf}, which takes a similar approach to our work by integrating the underwater image formation model \cite{akkaynak2019seathru} into a neural radiance field framework, specifically Mip-NeRF-360\cite{barron2022mipnerf360}. While SeaThru-NeRF estimates water parameters per viewing direction with a learned MLP, \Method assumes the water parameters are consistent for the whole scene and does not require such dense sampling. We also compare our method against vanilla 3D Gaussian Splatting \cite{kerbl3Dgaussians}. For both methods, we use the publicly released code from the authors.

\paragraph{Evaluation Metrics}
As we do not have ground truth underlying true color of the underwater scenes, unobtainable without draining the ocean, we present quantitative results on novel-view synthesis of the with-medium imagery. We evaluate visual fidelity by comparing rendering at held-out test frames to ground truth frames by computing peak signal-to-noise ratio (PSNR), structural similarity index (SSIM) \cite{wang2004imagessim}, and perceptual distance (LPIPS) \cite{zhang2018unreasonable}. We also evaluate computation requirements, time and memory, required both for training the learned representation and rendering a frame on a consistent set of hardware. 

\subsection{Results}

\begin{table*}[htbp]
\centering

\small

\resizebox{\textwidth}{!}{

\begin{tabular}{l|rrr|rrr|rrr|rrr|rrr|rrr|rrr}
\multicolumn{1}{l|}{} & \multicolumn{3}{c|}{Curaçao} & \multicolumn{3}{c|}{Japanese Gardens} & \multicolumn{3}{c}{Panama} & \multicolumn{3}{c}{IUI3} & \multicolumn{3}{c}{SimFog} & \multicolumn{3}{c}{SimWater} & \multicolumn{3}{c}{SaltPond} \\
& PSNR  & SSIM  & LPIPS & PSNR  & SSIM  & LPIPS & PSNR  & SSIM  & LPIPS & PSNR  & SSIM  & LPIPS & PSNR  & SSIM  & LPIPS & PSNR  & SSIM  & LPIPS & PSNR  & SSIM  & LPIPS \\                           
\midrule                                                                                                                                                                                                         
Ours                                    & \textbf{30.30} & \textbf{0.90} & \textbf{0.19} & \textbf{22.70} & \textbf{0.87} & \textbf{0.18} & 28.76 & \textbf{0.90} & \textbf{0.15} & \textbf{26.67} & \textbf{0.87} & \textbf{0.21} & \textbf{26.89} & 0.83 & 0.21 & \textbf{28.98} & \textbf{0.84} & 0.19 & \textbf{27.47} & \textbf{0.75} & \textbf{0.25} \\
 STN & 30.08 & 0.87 & 0.27 & 21.74 & 0.77 & 0.29 & 27.69 & 0.83 & 0.28 & 26.01 & 0.79 & 0.32 & 15.62 & 0.40 & 0.63 & 13.70 & 0.33 & 0.68 & 11.93 & 0.51 & 0.58 \\
 3DGS  & 28.01 & 0.88 & 0.21 & 21.47 & 0.85 & 0.22 & \textbf{29.64} & \textbf{0.90} & 0.17 & 21.11 & 0.81 & 0.29 & 26.38 & \textbf{0.84} & \textbf{0.20} & 28.90 & 0.85 & \textbf{0.18} & 27.10 & \textbf{0.75} & 0.29
 \\

\end{tabular}

}

\caption{
\textbf{Quantitative comparisons.} Across all datasets, \Method achieves competitive in-medium novel view rendering results. Compared to SeaThru-NeRF \cite{levy2023seathrunerf}, \Method achieves increased performance across all metrics on all datasets. Compared to 3D Gaussian Splatting \cite{kerbl3Dgaussians}, \Method achieves increased or similar performance on all metrics.  We \textbf{bold first-place} results. STN refers to SeaThru-NeRF \cite{levy2023seathrunerf}. Higher is better for PSNR and SSIM while lower is better for LPIPS (e.g. PSNR $\uparrow$, SSIM $\uparrow$, LPIPS $\downarrow$).
}
\vspace{-15pt}
\label{table:uw-comparisons}
\end{table*}


We report quantitative results on novel view synthesis of the original, in-medium images in \Cref{table:uw-comparisons}. We show increased performance as compared to SeaThru-NeRF \cite{levy2023seathrunerf} across all metrics and all datasets. Compared to 3DGS \cite{kerbl3Dgaussians}, we show increased or similar performance on all metrics. This suggests that we achieve high quality rendering of novel viewpoints, similar to or better than 3DGS \cite{kerbl3Dgaussians}, while also outperforming other methods \cite{levy2023seathrunerf} that similarly estimate the underlying color of the underwater scene. 

\begin{figure}[ht]
    \centering
    \includegraphics[width=\columnwidth]{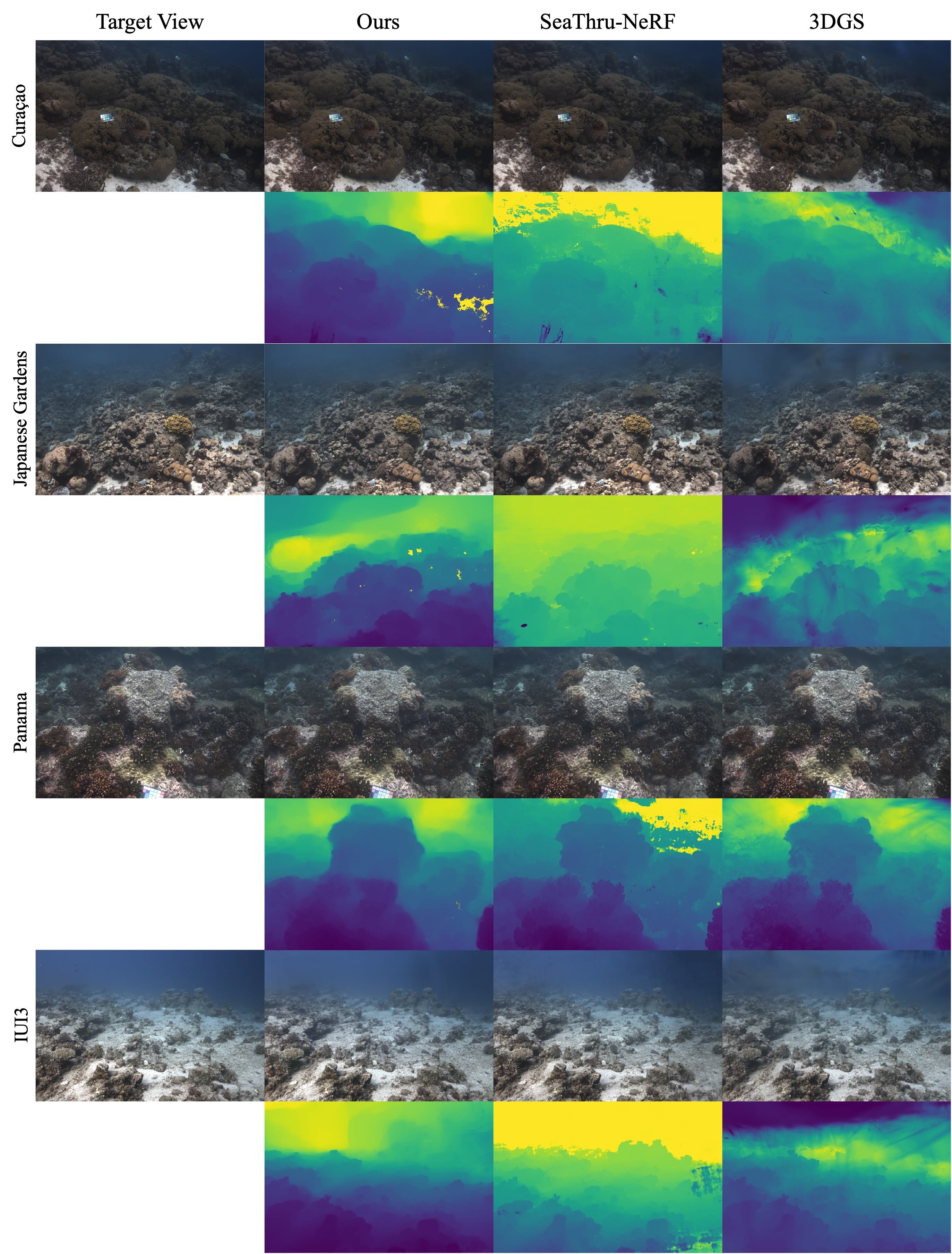}
    \caption{\textbf{Novel view synthesis with medium.} \Method produces high quality renders of unseen viewpoints similar to other state-of-the-art methods, SeaThru-NeRF \cite{levy2023seathrunerf} based on MiP-NeRF-360 \cite{barron2022mipnerf360} and 3D Gaussian Splatting (3DGS) \cite{kerbl3Dgaussians}. However, the depth renders are of noticeably higher quality than 3DGS which exhibit 3D Gaussian artifacts as well as floaters, as shown through the collapsed depth, representing the water column. Compared to SeaThru-NeRF, the depth renders are smoother and more coherent.}    
    \label{fig:nvs-with-depth}
    \vspace{-0.5cm}
\end{figure}

\begin{figure}[ht]
    \centering
    \includegraphics[width=0.9\columnwidth]{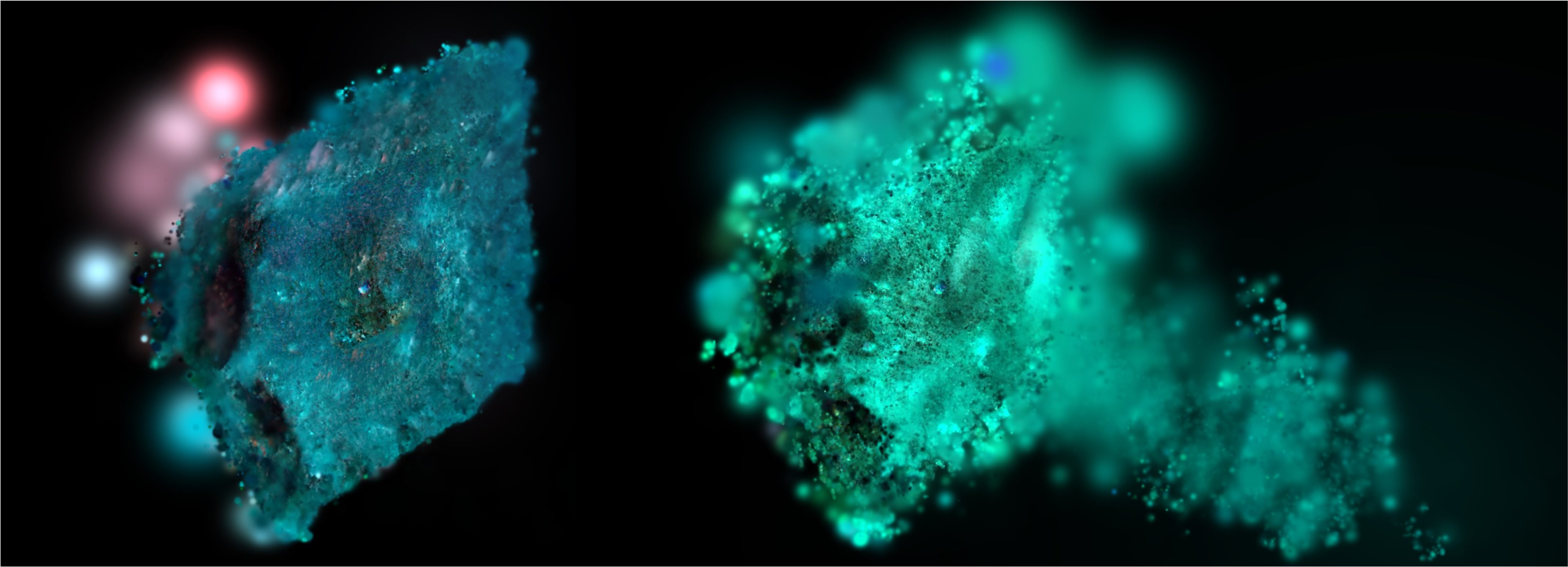}
    \caption{\textbf{3D Gaussian representation learned by \Method (left) and 3DGS \cite{kerbl3Dgaussians} (right).} Accounting for medium effects enables better learning of the underlying 3D scene structure as visualized on the Salt Pond downward-facing scene with the camera moving vertically in the water column. We see noticeably fewer floaters with \Method as compared to 3DGS.}    
    \label{fig:floaters-saltpond}
    \vspace{-0.3cm}
\end{figure}

Qualitatively, we visualize in-medium novel view synthesis results in \cref{fig:nvs-with-depth}. Looking at the color images, we see comparable results across all methods. However, examining the rendered depth images, we see much more variation. Notably, 3DGS tends to place 3D Gaussians in regions of low texture, specifically representing the water column, close to the camera. This effect can be seen in the depth image renders in top right region of Curaçao, top left of Japanese Gardens, top right of Panama, and top of IUI3. These floater regions lead to photometrically plausible renders but would be problematic from more extreme viewpoints. Like SeaThru-NeRF, \Method is able to infer a reasonable estimate of depth in these scenes with the extra constraints imposed by the underwater image formation model. We see smoother depth with \Method compared to 3DGS, which has many sharp, noisy artifacts, and compared to SeaThru-NeRF which exhibits small periodic artifacts in some regions. 

\begin{table}[ht]
\centering

\footnotesize


\begin{tabular}{l|rr|rr}
\multicolumn{1}{l|}{}& \multicolumn{2}{c|}{Inference Time} & \multicolumn{1}{c}{} \\
& Train $\downarrow$ & Render $\downarrow$ & VRAM (GB) $\downarrow$  \\                           
\midrule                                                                                                                                                                                                          
 Ours                                   & 1 h 25 m  & 0.012 s  & 4.0   \\
 Levy et al.~\cite{levy2023seathrunerf} & 21 h    & 10.184 s   & 33.2  \\
 Kerbl et al. ~\cite{kerbl3Dgaussians}  & 40 m    & 0.006 s    & 3.8   \\

\end{tabular}


\caption{
\textbf{Computational constraint comparisons.} \Method preserves the computational efficiency of 3D Gaussian Splatting \cite{kerbl3Dgaussians}, requiring significantly lower train time and memory constraints while still enabling real-time rendering than other underwater restoration methods.}
\vspace{-15pt}
\label{table:computational-comparison}
\end{table}

\begin{table}[ht]
\centering

\footnotesize


\begin{tabular}{l|rrr}
\multicolumn{1}{l|}{} & \multicolumn{3}{c}{SeaThru-NeRF Dataset} \\
& PSNR $\uparrow$ & SSIM $\uparrow$ & LPIPS $\downarrow$  \\                           
\midrule                                                                                                                                                                                                         
Vanilla 3DGS            & 24.42 & 0.85 & 0.25 \\
+ DS                    & 24.03 & 0.84 & 0.25 \\
+ C                     & 24.18 & 0.84 & 0.26 \\
+ BG                    & 26.84 & 0.88 & 0.20 \\
+ BS                    & 23.91 & 0.83 & 0.26 \\
+ DS + BG               & 26.71 & 0.87 & 0.19 \\
+ DS + BG + C           & \textbf{27.13} & 0.88 & \textbf{0.18} \\
+ DS + BG + C + BS      & 26.64 & 0.88 & 0.19 \\
\textbf{Ours (all losses)}     & 27.11 & \textbf{0.89} & \textbf{0.18} \\

\end{tabular}


\caption{
\textbf{Ablations.} We ablate the various components of our loss function, measuring performance averaged across the SeaThru-NeRF datasets for the task of in-medium novel view synthesis. SD refers to smooth depth loss, C refers to the color losses (gray world and saturation), BS refers to backscatter loss, and BG refers to background loss. 
}

\vspace{-15pt}
\label{table:ablations}
\end{table}

The difference in depth estimation between \Method and 3DGS is most clearly visualized on the Salt Pond dataset, where the camera trajectory is solely up and down within the water column with downward facing imagery. The 3D Gaussian representation learned by each method is visualized from an extreme viewpoint in \cref{fig:floaters-saltpond} (\Method on the left and 3DGS on the right). While \Method learns a representation consisting of solely the seafloor, 3DGS places many floaters within the water column to create the degraded image quality, a result of the medium, present in the original input images.

\begin{figure*}[htbp]
    \centering
    \includegraphics[width=0.9\linewidth]{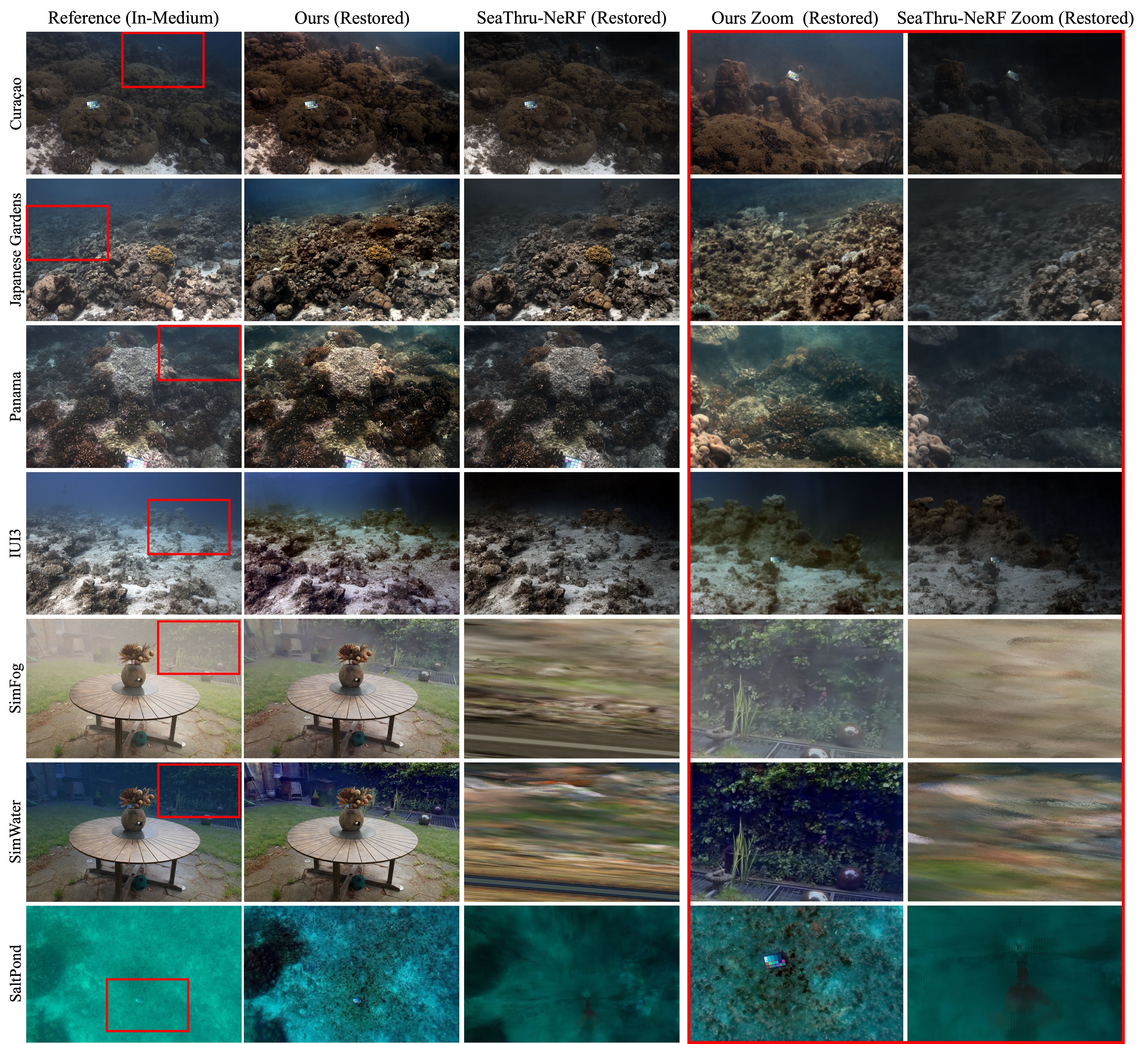}
    \caption{\textbf{Novel view synthesis restoring underlying color.} \Method produces a more vivid and qualitatively pleasing restoration of the color of the scene. In particular, details in the foreground are brought out more clearly, especially in contrast to SeaThru-NeRF \cite{levy2023seathrunerf} which produces are more muted and less vivid rendering of distant parts of the scene. On the far left column, we show the captured reference image in the medium.  We show the true color, J, of both the full image as well as a zoomed-in crop of a more distant portion of the scene. The cropped portion is highlighted by the red box in the reference image. Note that SeaThru-Nerf \cite{levy2023seathrunerf} performs poorly with the synthetic dataset as well as the robot collected SaltPond dataset.}
    \label{fig:qual-truecolor}
    \vspace{-0.5cm}
\end{figure*}


To understand the abilities of \Method compared to SeaThru-NeRF in restoring the true color of the scene, we visualize the inferred restorations in \Cref{fig:qual-truecolor}. Across the SeaThru-NeRF datasets, appearance look reasonable with both methods, though \Method shows some brighter and more vivid restorations. Examining regions attenuated and covered by the veiling effect, we see much more recovered detail and color. Looking at the synthetically added fog and water onto the Garden scene, \Method is able to mitigate the effects of fog and water, allowing for more detail and color to be perceived with the flora in the background, while SeaThru-NeRF struggles to recover a reasonable estimate of the scene. On the Salt Pond dataset, which is a downward facing, bounded scene, we highlight the ability of \Method to enhance the image, as shown through the color chart being much clearer even at such a high altitude. 


We also show a comparison in computational requirements between different methods in \Cref{table:computational-comparison}, averaged across training and evaluation on the SeaThru-NeRF datasets. Our method has minimal additional inference time and memory constraints as 3D Gaussian Splatting, showing our additional medium estimation procedures to not negligibly affect performance. The increase in rendering time can largely be attributed to a second rendering pass required to obtain depth. This preserves and reinforces the significant paradigm improvements from NeRFs to 3D Gaussians as seen in the high computational requirements for Sea-Thru NeRF. Furthermore our method does not require densely querying or sampling medium parameters at every pixel, instead having a set of global medium parameters. 

\subsection{Ablations}

Finally in \cref{table:ablations}, we also ablate design decisions - namely, the role of the additional loss objectives we've added. Note that these ablations quantify the performance on in-medium novel view synthesis while we are also concerned with the task of color restoration. We can see that the background loss is helpful while many of the other loss components individually do not lead to significant gains. However, when combined together we see that our total loss function does lead to increased performance.


\section{Conclusion}
\label{sec:conclusion}

We present \Method, a method that enables real-time rendering of underwater scenes combining advances in 3D radiance fields with a physically-grounded underwater image formation model. On a range of real-world and simulated datasets, \Method generates high quality novel-views in medium while also restoring the true, underlying color of these scenes without the medium. At the same time, \Method adds minimal computational overhead to 3D Gaussian Splatting, preserving the real-time rendering capabilities, and significantly outperforming other radiance field methods that do restore the color against effects of a degrading medium. 

\textbf{Limitations and Future Work} Before deployment on any underwater robot, the method needs to be adapted to run in real-time environments, building out the 3D Gaussian representation as new data is observed. This opens up potential uses in underwater navigation, collision avoidance, and adaptive sampling of visually interesting phenomena. In terms of fully modelling the effects of light underwater, this work does not consider interactions between light and the surface, (e.g., caustics) as done in other recent work \cite{zhang2024recgs} and as well as the shadows in the scene resulting from divers or an underwater robot, as are present in the Salt Pond dataset. Finally, the underwater world is filled with dynamic, moving objects (e.g. fan coral that sway with the currents) and our work focuses on static, underwater environments.

\section*{ACKNOWLEDGMENTS}
This work was supported in part by The Investment in Science Fund at WHOI, and NSF OCE Award No. 2333604.

{\small
\bibliographystyle{IEEEtran}
\bibliography{egbib}
}

\end{document}